# Towards Offensive Language Identification for Tamil Code-Mixed YouTube Comments and Posts

Charangan Vasantharajan · Uthayasanker Thayasivam

**Abstract** Offensive Language detection in social media platforms has been an active field of research over the past years. In non-native English spoken countries, social media users mostly use a code-mixed form of text in their posts/comments. This poses several challenges in the offensive content identification tasks, and considering the low resources available for Tamil, the task becomes much harder. The current study presents extensive experiments using multiple deep learning, and transfer learning models to detect offensive content on YouTube. We propose a novel and flexible approach of selective translation and transliteration techniques to reap better results from fine-tuning and ensembling multilingual transformer networks like BERT, DistilBERT, and XLM-RoBERTa. The experimental results showed that ULMFiT is the best model for this task. The best performing models were ULMFiT and mBERT-BiLSTM for this Tamil code-mix dataset instead of more popular transfer learning models such as DistilBERT and XLM-RoBERTa and hybrid deep learning models. The proposed model ULMFiT and mBERT-BiLSTM yielded good results and are promising for effective offensive speech identification in low-resourced languages.

Charangan Vasantharajan*
Department of Computer Science and Engineering, University of Moratuwa, Sri Lanka
*charangan.18@cse.mrt.ac.lk*

Uthayasanker Thayasivam*
Department of Computer Science and Engineering, University of Moratuwa, Sri Lanka
*rtuthaya@cse.mrt.ac.lk*

# 1 Introduction

Offensive speech is commonly defined as any communication that causes an individual or a group to feel upset, resentful, annoyed, or insulted based on the characteristics such as color, ethnicity, sexual orientation, nationality, race, and religion. These type of talks lasts forever on these social media platforms such as Facebook, Twitter, YouTube and so on, compared to physical abuse and affects the individual on the mental status creating depression, insomnia, and even suicide. Therefore, the need for identifying and restraining hatred speech is essential. But, detecting such content is a very critical issue due to the enormous volume of user-generated multilingual data on the web, particularly social media platforms [1].

In the past few years, with a proliferation of social media users from the multilingual society, posts and comments are delivered in a **code-mixed** form. What is Code-Mixed? A sentence that contains mixed vocabulary and syntax of multiple languages [2]. Due to the lack of impressions in a single language, most users use code-mixed text to convey their thoughts with a multi-linguistic community[3]. Moreover, users tend to use roman characters for texting instead of the native script is also identified. This leads to severe challenges for offensive speech identification by the lack of methodologies for handling code-mixed and romanized text[4].

Due to lack of moderation, a significant amount of offensive content in the form of code-mixed is often posted on these platforms. Since social media has become an indispensable part of our life, the content posted on social media has a great impact on society. It has been revealed that offensive and provoking content on these platforms can lead to riots. Hence the moderation of content in these platforms is a very important task to purify social media platforms but also conductive to promote positive development in the society[5]. It is impossible to identify such content manually from the large amounts of data being generated every moment on social media, so there is a need to perform automated moderation of the content. This has encouraged a significant number of researchers who are in the domain of Natural Language Processing(NLP) to develop robust computational systems to limit the spread of offensive content using the state-of-the-art NLP techniques[6].

The goal of our current study is to classify offensive language from a Tamil code-mixed dataset of comments and posts collected from YouTube[7]. The comments/posts contain multiple types of offensive content in the dataset. For this purpose, a multi-class classification method is to be adapted to our task. Recent studies in multilingual text classification are led by transformer-based networks like BERT [8] and XLM-RoBERTa [9]. An additional advantage of these architectures is that it yields good results even with the low-resourced languages. This aspect is beneficial to Tamil due to not having properly established datasets. To accomplish our task, we examined a few transfer-learning models like BERT, DistilBERT, and XLM-RoBERTa as well as a deep learning model CNN-BiLSTM. However, there is a caveat directly using the models on the romanized or code-mixed text: the transformer models are trained on languages in their native script, not in the romanized script in which users prefer to write online. We solve this issue by using a novel way to convert such sentences into their native language while preserving their semantic meaning - selective translation and transliteration. Our important contributions are as follows.

- Proposed selective translation and transliteration for text conversion in romanized and code-mixed settings which can be extended to other romanized and code-mixed contexts in any language.
- Experimented and analyzed the effectiveness of fine-tuning and the ensemble of transformer-based models for offensive language identification in code-mixed and romanized scripts.

The rest of the article is organized as follows, Section 2 presents an overview of the articles proposed on Offensive Language Detection. Section 3 entails a detailed description of the task and dataset. This is followed by the explanation of proposed models in Section 4. The experiments and results are explained in Section 5 and Section 6. Section 7 analyzes our results by using a confusion matrix. The paper concludes by comparing and highlighting the main findings with a benchmark.

## 2 Related Work

Offensive content in social media might trigger objectionable consequences to its user like mental health problems, and suicide attempts [1]. To keep the social media ecosystem, coherent researchers and stakeholders should try to develop computational models to identify and classify offensive content and restrain it within a short period. So, there is a lot of research has already been done.

In the early stages, computational models were created by using support vector machine(SVM), Naive Bayes, and other traditional machine learning approaches [10]. [11] investigated which can be effect more on pretrained models BERT and Machine Learning Approaches for Offensive Language Identification in English, Danish and Greek languages. They trained the models using Logistic Regression, Multinomial Naive Bayes, Random Forest and Linear Support Vector Machine. BERT model and Logistic Regression performed well in their task. [12] addressed all code-mixed and imbalance distribution related issues and came up with a solution to analyze sentiments for a class imbalanced code-mixed data using sampling technique combined with levenshtein distance metrics. Furthermore, this paper compared the performances of different machine learning algorithms namely, Random Forest Classifier, Logistic Regression, XGBoost classifier, Support Vector Machine and Naïve Bayes Classifier using F1- Score.

[13] explored in Kannada-English by the traditional learning approaches such as Logistic Regression(LR), Support Vector Machine(SVM), Multinomial Naive Bayes, K-Nearest Neighbors(KNN), Decision Trees(DT), and Random Forest (RF). The performance of these models has not reached a state of the art as they could not detect the semantic and contextual relationship in texts. This problem was mitigated with the arrival of word embeddings and Recurrent Neural Networks(RNN) [14]. On the other hand, to encourage sentiment analysis tasks and overcome the non-availability of the code-mixed dataset in Tamil-English, [2] created an annotated corpus containing 15744 comments and posts from YouTube. In addition to this, they examined a benchmark system with various machine learning algorithms such as Logistic Regression(LR), Support Vector Machine(SVM), K-Nearest Neighbour(K-NN), Decision Tree(DT), Random Forest(RF), Multinominal Naive Bayes, 1DConv-LSTM, and BERT Multilingual. In this task, the Logistic regression, random forest classifiers, and decision trees were provided better results compared to other algorithms. Moreover, [15] introduced a methodology to detect Sinhala and English words in code-mixed data by presenting a language detection

model with XGB classifier. [16] created a newly code-mixed dataset with sentence and word-level annotation for Sinhala-English language pairs collected from Facebook comments, chat history, and public posts.

According to the social media explosion, [17] performed an analysis to separate the general profanity from the hate speech in social media. They presented a supervised classification system that used character n-grams, word n-grams, and word skip grams. Their model was able to achieve 80% accuracy on a huge dataset which contained English tweets annotated with three labels (hate speech (HATE), offensive language but no hate speech (OFFENSIVE), and no offensive content (OK)). [18] addressed the problem of offensive language identification by creating an annotated corpus using comments retrieved from Twitter Posts for the Greek language. They experimented with seven different classification models: GRU with Capsule [19], and BERT [8]. In this task, LSTM and GRU with Attention performed well compared to all other models in terms of macro-f1, because bidirectional LSTMs and GRUs can hold contextual information from both past and future, creating more robust classification systems[20]. However, the fine-tuning approach with the BERT-Base Multilingual Cased model did not provide good results.

In recent years, transformer-based model such as BERT[8], DistilBERT[21], XLM-RoBERTa[9] gained more attention to identify and classify offensive texts through contextual and semantic learning. These large pre-trained models can classify code-mixed texts of different languages with astonishing accuracy[2]. [22] examined transformer networks to predict named entities in a corpus written both in native as well as in Roman script for English-Hindi language pairs. Based on the above analysis, most of the approaches were unidirectional-based except the works done by [2] and [18]. They used BERT [8] as their pre-trained model but it did not give state-of-the-art results on code-mixed text. After a deep dive into previous studies and observations about recent developments, we chose transformer-based models and a deep learning-based model for this low-resource language experiment as an extension work of [23].

## 3 Datasets

The Tamil-English dataset used in this work was taken from a competition - Offensive Language Identification in Dravidian Languages [7] (henceforth ODL). The objective of the ODL task was a multi-class classification problem with multiple offensive categories (Not-Offensive, Offensive-Untargeted, Offensive-Targeted-Insult-Individual, Offensive-Targeted-Insult-Group, Offensive-Targeted-Insult-Other, Not-Tamil).

The dataset provided is scrapped entirely from the YouTube posts/comments of a multilingual community where code-mixing is a prevalent phenomenon. This derived dataset comprised of all six types of code-mixed sentences: No code-mixing, Inter-sentential Code-Mixing, Only Tamil (written in Latin script), Code-switching at morphological level (written in both Latin and Tamil script), Intra-sentential mix of English and Tamil (written in Latin script only) and Inter-sentential and Intra-sentential mix. To understand the statistics of misspelled and code-mixed words, we compared with an existing pure language vocabulary available in the **Dakshina** dataset [24]. We find the proportion of the out-of-vocabulary (OOV) words (including code-mixed, English, and misspelled words) in the dataset as 85.55%.

The dataset consists of comments in six different classes:

1. **Not-Offensive**: Comments which are not offensive, impolite, rude, or profane.
   e.g.: I am waiting vera level trailer i ever seen
2. **Offensive-Targeted-Insult-Individual**: offensive comments targeting an individual.
   e.g.: Sarkar Teaser likes ah beat pana mudilayeda... Ama payale 🤣🤣🤣🤣 Iththu ponavane
3. **Offensive-Targeted-Insult-Group**: offensive comments targeting a group.
   e.g.: Wig mandaiyan fans,vaaya mudikittu irukkanum
4. **Offensive-Targeted-Insult-Other**: offensive comments targeting an issue, an organization, or an event other than the previous two categories.
   e.g.: Inimae ajith fans evanadhu.... Copy adikuradha pathi pesuvenga...
5. **Offensive-Untargeted**: offensive comments targeting no one.
   e.g.: Kola gaandula irruke mavane kollama vida maatan
6. **Not-Tamil**: comments not in Tamil.
   e.g.: Hindi me kab release hoga plz comment

We present the dataset statistics in table 1 and visualize in Figure 1.

## 4 System Description

### 4.1 Preprocessing

The dataset described in the section 3, needs some pre-processing to be able to adapt to machine learning algorithms. The pre-processing used in this paper are as follows:

| Label | Train | Dev | Test |
|---|---|---|---|
| Not-Offensive | 25425 | 3193 | 3190 |
| Offensive-Targeted-Insult-Individual | 2343 | 307 | 315 |
| Offensive-Targeted-Insult-Group | 2557 | 295 | 288 |
| Offensive-Targeted-Insult-Other | 454 | 65 | 71 |
| Offensive-Untargeted | 2906 | 356 | 368 |
| Not-Tamil | 1454 | 172 | 160 |
| Total | 35139 | 4388 | 4392 |

Table 1: Number of instances for each class in train, validation and test sets. The imbalance of the dataset depicts a realistic picture observed on social media platforms.

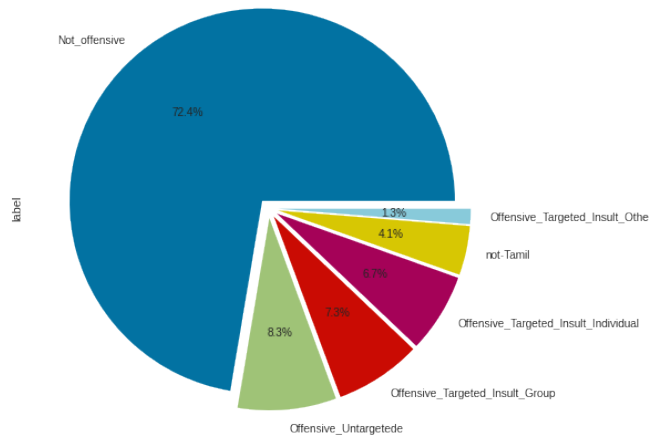

Figure 1: Class distribution on Training set. Dataset is highly imbalanced where a number of instances in Not-Offensive class is much higher compare to other classes.

– Firstly, the words in the script are employed to Lemmatization, Stemming, removing snip words to preserve context to the users' intent, and Lower casing the Romanized words. This step is not performed for words written in Tamil script as there is no such casing used in Tamil script.
– Secondly, we remove emojis and emoticons from text, as well as remove all special characters, numbers, punctuation, and user mentions as they generally do not carry any semantic meaning.

Further, since we use transformer models (contextual models), it is observed that stop words receive a similar amount of attention as non-stop words, so we did not remove stop words. We provide some examples for the above techniques below.

1. Inda music ha engaiyoo keta mariyae irukae?? 🤔 🤔
   Preprocessed: Inda music ha engaiyoo keta mariyae irukae
2. அட அரைவேக்காட்டு பயலே :-D :-D
   Preprocessed: அட அரைவேக்காட்டு பயலே
3. #6Million Views @5Days..!! #200K Came In Quick Baby..!! #Varlaam #Varalaaam Vaa #Bairavaa...!!!
   Preprocessed: Views Came In Quick Baby Vaa
4. ThalaivanSTR <3 #Vjs <3 #AV <3 #AS <3!
   Preprocessed: ThalaivanSTR

### 4.2 Selective translation and transliteration(STT)

This is a novel idea we have used to get a proper representation of text in the native script for the final neural architecture training. The pseudo-code for the proposed algorithm is given in Algorithm 1. The primary need for this step is as follows: Recent advancements in state-of-the-art multilingual NLP tasks are led by Transformer architectures like mBERT and DistilmBERT which are trained on multiple languages in the native script but not in the romanized script. Hence to reap better results by fine-tuning these architectures, the text is to be in a native script (for example, Tamil text in Tamil script).

To convert text into the native script, we cannot rely on neural translation systems, particularly in tweets where users tend to write informally using multiple languages. Also, translating romanized non-English language words into that language does not make any sense in our context. For example, translating the word

"Valthukkal"(which means congrats in English) directly into Tamil would seriously affect the entire sentence's semantics in which the word is present because "Valthukkal" will be treated as an English word. In many cases, valid translations from English to a non-English language would not be available for words.

So, as a solution to this problem, we propose selective transliteration and translation of the text.

In this conversion pipeline, we checked the word in three steps as follows and applied rel event functions calls.

1. If the word is in English dictionary, then we translated the word into native Tamil script

2. We ignored the word if the detected language is native Tamil

3. If the word not processed by the above both cases, then we transliterated the romanized word into native Tamil.

The segregation of English words and native language words in each sentence is done using a big corpus of English words from NLTK-corpus[1]. The idea of this selective conversion is based on the observation that in romanized native language comments(like Tanglish), users tend to use English words only when they can convey the meaning better with the English word or when the corresponding native language word is not much used in regular conversations. For example, the word "movie" is preferred by Tamil users over its corresponding Tamil word.

The translation of words is done using Google Translate API[2], and transliteration is done with the help of AI4Bharat Transliteration[3]. The detection of language script is carried out with a help of NLTK words.

In some tweets, the direct translation of intermediate English words present in them can affect the semantics negatively, and it may be better to keep them as they are. So, we have experimented with this variation of the STT algorithm - only transliterate the non-English words into their corresponding language and keep the English words as they are. We observed that the cross-lingual nature of the transformer-based models can efficiently handle these types of mixed script sentences.

1. Older me —- engineer aganum miss Me know —— Yedhuku engineering padicheyno

---

[1] https://www.nltk.org/
[2] https://pypi.org/project/googletrans/
[3] https://pypi.org/project/ai4bharat-transliteration/

**Algorithm 1** Algorithm for Selective Translation and Transliteration of code mixed and romanized languages

**Input**: romanized or Code-mixed text T and desired native language for the final script L

**Output** Text in native script

1: **procedure** STT(*T,L*)  ▷ This is the STT
2:  Initialization: EngWords = Set of all English words
3:  words = splitLineIntoWords(T)
4:  LOOP Process
5:  **for** $i \leftarrow 0$ to *len*(*words*) **do**
6:   word = words[i]
7:   **if** *detect*(*word*) = *L* **then**
8:    continue
9:   **else if** *wordinEngWords* **then**
10:    words[i] = translate(word, L)
11:   **else**
12:    words[i] = transliterate(word, L)
13:  **return** *joinWords*(*words*)

STT Output: Ý ñ ä ß Í Ü õ ñ Ü ï Ý ê á ë ß ê ã à õ Ç Ó$y Þ õ ï â õ å ë Í Ü õ ñ Ü ï Ú à ë$r Þ õ Í$n$h ï Ý ê á ë ß ë ß â õ Ý'2 ï

2. apadiye kaathukku panchachayum kodungada.........kaathu poora raththam

STT Output: Ç Ý õ Ý'2 ð ß Ó ê$n Ó õ$h Ý × õ Õ ð Õ$r Þ õ ï Ó Ø ê Ó ê$n $, à ê à Ú õ Ú Þ õ

### 4.3 Models

Recent studies show that pre-trained word embeddings and fine-tuning of state-of-the-art Transformer architectures show better performance in text classification compared to classical machine learning approaches like N-gram features with bag of words models [25]. So, we directed our entire focus on using the word embeddings of transformer architectures both pre-trained and finetuned for text classification along with CNN-BiLSTM. The text obtained using selective translation and transliteration is used in further steps.

#### 4.3.1 CNN-BiLSTM

This is a hybrid of bidirectional LSTM and CNN architectures that extracts a new feature vector from the per-character feature vectors for each word by employing a convolution and a max-pooling layer. For each word, these vectors are concatenated and fed to the BiLSTM network and then to the output layers. CNN-BiLSTM, along with Doc2Vec embedding achieved very high results for sequence classification tasks [26], thus we use GLoVe[4]. embedding.

---

[4] https://nlp.stanford.edu/projects/glove/

### 4.3.2 mBERT-BiLSTM

Multilingual models of BERT (mBERT) are largely based on the BERT architecture [8] pre-trained on the Wikipedia dump of top 100 languages using the same training strategy that was employed to BERT, i.e., Masked Language Modeling (MLM) and Next Sentence Prediction (NSP). To account for the data imbalance due to the size of Wikipedia for a given language, exponentially smoothed weighting of data was performed during pre-training data creation and word piece vocabulary creation. This results in high resource languages being under-sampled, while low resourced languages being over-sampled.

### 4.3.3 DistilBERT

DistilBERT [21] follows the same architecture of that of BERT [8], while reducing the number of layers by a factor of 2. DistilBERT follows a triple loss language modeling, which combines cosine distance loss with knowledge distillation for it (student) to learn from the larger pre-trained natural language model (teacher) during pretraining. It has 40% fewer parameters than BERT, runs 60% faster while preserving over 95% of BERT's performances as measured on the GLUE language understanding benchmark[5].

### 4.3.4 XLM-RoBERTa

XLM-RoBERTa [9] is a large multi-lingual language model, trained on 2.5TB of cleaned Common Crawled data in 100 languages. It can be recognized as a union of XLM [27] and RoBERTa [28]. The training process involves sampling streams of text from different languages and masking some tokens, such that the model predicts the missing tokens. Using SentencePiece [29] with a unigram language model [30] subword tokenization is directly applied on raw text data. Since there are no language embeddings used, this allows the model to better deal with code-switching. XLM-RoBERTa manifested remarkable performance in various multilingual NLP tasks.

### 4.3.5 ULMFiT

ULMFiT stands for Universal Language Model Fine-tuning for Text Classification and is a transfer learning technique that involves creating a Language Model that can predict the next word in a sentence, based on unsupervised learning of the WikiText 103 corpus.

The ULMFiT model uses multiple LSTM layers, with dropout applied to every layer and developed as the AWD-LSTM architecture [31]. We prepared our dataset in the format that FastAI[6] requires it to be in. FastAI provides simple functions to create Language Model and Classification "data bunch". Creating a data bunch automatically results in pre-processing of text, including vocabulary formation and tokenization. We created a language model with the AWD-LSTM model. Then we use transfer learning to update the model to train on a specific dataset that may be quite small. Final stage is the classifier fine-tuning which preserves the low-level representation and updates the high-level weights using gradual unfreezing of the layers.

## 5 Experiment Setup

All our models are culled from **HuggingFace**[7] transformers library and the models' parameters are as stated in Table 2. The architecture of deep learning and transformer-based models are depicted in Figure 2 and Figure 3 respectively.

| Hyperparameter | Value |
|---|---|
| LSTM Units | 256 |
| Dropout | 0.2 |
| Activation Function | Softmax |
| Max Len | 128 |
| Batch Size | 64 |
| Optimizer | AdamW |
| Learning Rate | 1e-5 |
| Loss Function | Cross-Entropy |
| Epochs | 10 |

Table 2: Common parameters for the models that we used during our experiments.

### 5.1 CNN-BiLSTM

In our CNN model(Figure 2), the first layer is the GloVe[8] Embedding layer that uses 100-dimension vectors to represent each word. The next layer is the bidirectional LSTM layer followed by the 1D convolutional layer. Then the output of the Con1D layer is fed into a max-pooling layer, and dropout for regularization, respectively. Finally, because this is a classification problem, we use a Dense output layer with six neurons and a SoftMax activation function to make predictions for the six classes in the problem.

---

[5] https://huggingface.co/transformers/model_doc/distilbert.html
[6] https://docs.fast.ai/
[7] https://github.com/huggingface/
[8] http://nlp.stanford.edu/data/glove.6B.zip

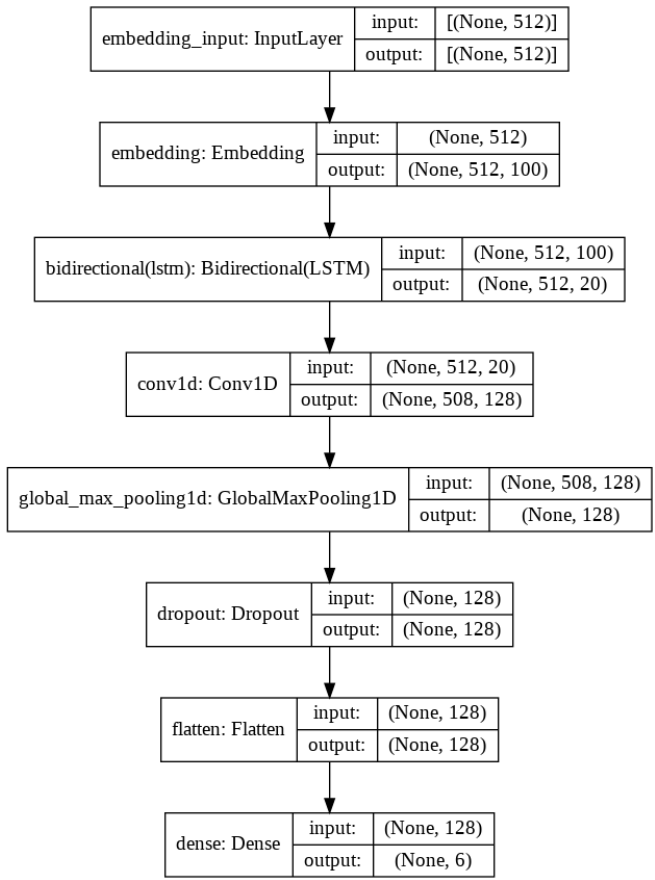

Figure 2: : Proposed hybrid CNN network involves using Convolutional Neural Network (CNN) layers for feature extraction on input data combined with bidirectional LSTMs to support sequence prediction.

### 5.2 mBERT-BiLSTM

**bert-base-multilingual-cased** is a pre-trained BERT multilingual model having approximately 110M parameters with 12-layers, 768 hidden-states, and 12-heads[9] was used. During the fine-tuning step, bidirectional LSTM layers were integrated into the model and then fed embeddings as the input to it which leads to the increase in the information being fed which results in the improvement of the context and precision.

### 5.3 DistilBERT

DistilBERT is a small, fast, cheap, and light Transformer model trained by distilling BERT base on the concatenation of Wikipedia in 104 different languages. We fine-tuned **distilbert base multilingual cased** model having 6 layers, 768 dimension, 12 heads, and total-izing 134M parameters (compared to 177M parameters for mBERT-base).

[9] https://github.com/google-research/bert

### 5.4 XLM-RoBERTa

We used **XLM-RoBERTa-base**, a pre-trained multilingual model that has been trained on more than 100 languages for our sequence classification. This model has 12 Layers, 768 Hidden, 12 attention heads, and 270M parameters and is trained on 12.2 GB of monolingual Tamil in addition to 300.8 GB of English corpus[9]. This allows the model for effective cross-lingual transfer learning. As we were primarily dealing with native Tamil script data, it was effective.

### 5.5 ULMFiT

We did more epochs after unfreezing all the layers of the pre-trained model. This process will train the whole neural network rather than just the last few layers. To train the classifier, we used a technique called gradual unfreezing. We started by training the last few layers, then went backward and unfreeze and trained layers before. This method involves fine-tuning a pre-trained language model (LM) AWD-LSTM to a new dataset in such a manner that it does not forget what it previously learned. AWD-LSTM language model has an embeddings size of 400 and 3 layers which consists of 1150 hidden activations per layer. It also has a batch size of 64 and is fine-tuned by adhering to the slanted triangular learning rates by freezing a few of the layers and dropouts with a multiplier of 0.5 were applied.

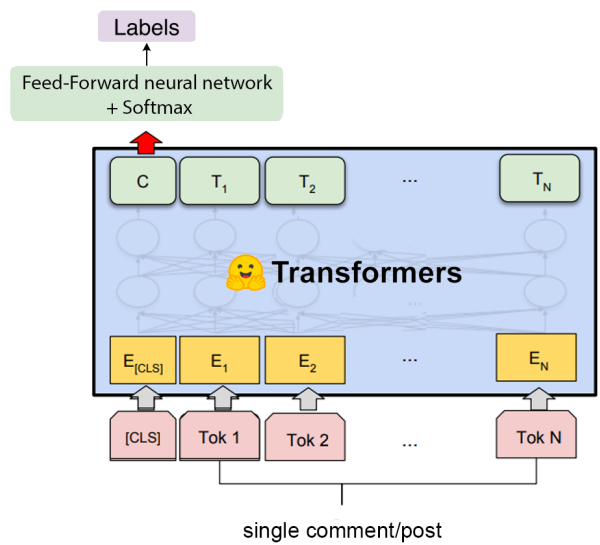

Figure 3: Transformer based Models' Architecture. To recreate this image, we used a source image from [8].

## 6 Results and Analysis

This section presents the experimental results of all three models explained in Section 5. The results with weighted average are presented in terms of precision, recall, and F1-score of class not-offensive, Offensive-Targeted-Insult-Individual, Offensive-Targeted-Insult-Group, Offensive-Targeted-Insult-Other, Offensive-Untargeted, Not-Tamil in Table 3. A particular model is identified as best if it has reported the highest weighted average of precision, recall, and F1-score. The results on the test set are tabulated in Table 3.

The evaluation metric of this task is weighted average F1-score due to account for the class imbalance in the dataset. Among the mentioned models, ULMFiT and mBERT-BiLSTM produced good weighted avg F1-scores of 0.7346 and 0.7341 on the test dataset, respectively. CNN-BiLSTM and XLM-RoBERTa models gave almost similar F1-scores of 0.6112 and 0.6111 respectively on the same dataset. We submitted DistilBERT model as it has obtained an F1-score of 0.6032, and it is slightly less than the F1-score of CNN-BiLSTM and XLM-RoBERTa. According to Table 3, the ULMFiT model has marginally outclassed mBERT-BiLSTM and other transformer-based models.

Models like CNN-BiLSTM, DistilBERT, and XLM-R gave similar and poor results on the test dataset. One of the reasons for the poor performance of these models is the imbalance in the distribution of the classes. In the dataset, most of the comments/posts belong to the class of not-offensive while the other classes like not-Tamil, offensive-targeted-insult-group, offensive-targeted-insult-other, offensive-untargetede have a small classification of texts. These models performed better on the majority class and poorly on the minority classes. XLM-RoBERTa gave better results on the validation set(72% accuracy), but due to the class imbalances, it could have underperformed on the test set. It is observed that the CNN-BiLSTM model also performed poorly.

| Models | Precision | Recall | F1-Score |
|---|---|---|---|
| ULMFiT | 0.7287 | 0.7682 | 0.7401 |
| mBERT-BiLSTM | 0.7182 | 0.7579 | 0.7341 |
| CNN-BiLSTM | 0.5741 | 0.6754 | 0.6112 |
| XLM-RoBERTa$_{base}$ | 0.5275 | 0.7263 | 0.6111 |
| DistilmBERT$_{cased}$ | 0.5146 | 0.7169 | 0.6032 |

Table 3: Weighted F1-scores according to the models on the test dataset.

In the CNN-BiLSTM model, the convolution layer was not capturing the correlations and patterns within the input. Moreover, the BiLSTM layer did not apprehend the dependencies within the attributes extracted by the CNN layer, which has led to the poor performance of the model. For the word embeddings, we used GloVe embedding which did not perform well on the CNN. Fine-tuning the transformer model Multilingual BERT-BiLSTM has resulted in good performance than DistilBERT. ULMFiT models attained a better performance in predicting the minority classes as well. The major reasons for the better performance of ULMFiT over other models are due to its superior fine-tuning methods and learning rate scheduler.

## 7 Error Analysis

Table 3 shows that ULMFiT and mBERT-BiLSTM are the best-performing languages models for Tamil. To get more insights, we present a detailed error analysis of these models carried out by using the confusion matrix (Figure 4).

In the view of Figure 4, only NF and NT classes have a high true-positive rate (TPR) while in other classes, texts have mostly been misclassified as not-offensive. However, among all six classes, none of the classes got 0% TPR on the other hand, our model identified only 1 OTIO class among 71 texts more than 60% wrongly classified as NF. The high-class imbalance situation might be the reason for this vulnerable outcome. This resulted in the rate of misclassification of OTIO class that is more than 98%.

Figure 5 reveals that the model is biased towards NF class. The model failed to identify none of the 71 OTIO texts. The insufficient number of instances have been confirmed with ULMFiT and mBERT. Moreover, among 288 OTIG texts, only 34 have been correctly classified. Meanwhile, we observe that only NF, OU, and NT classes have a higher TPR wherein most of the cases, the model misclassified as NF class (Figure 5) as same as in the Figure 4. In contrast, OTIO and OTII classes have shown misclassification rates of 100% and more than 99%.

In comparison to the transformer-based models, the deep learning model CNN-BiLSTM shows a slightly different confusion matrix (Figure 6). The model gave high TPR for only NF classes while in other classes, texts have mostly been misclassified as not-offensive. As same as in the 5, OTIO got 0% TPR, and predicted only 2 of 160 NT classes with more than 98% misclassification rate. Overall, a lower number of test datasets for OTIO class mostly led to 100% misclassification and we observed a class imbalance problem in

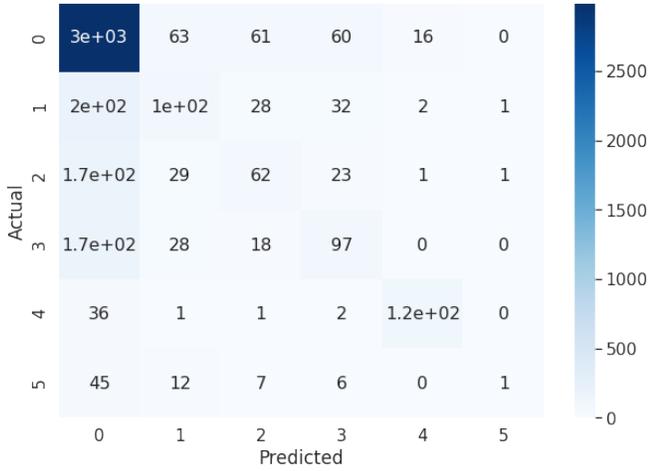

Figure 4: Confusion matrix of the best model - ULMFiT. 0-Not Offensive(NO), 1-Offensive Untargetede(OU), 2-Offensive Targeted Insult Group(OTIG), 3-Offensive Targeted Insult Individual(OTII), 4-Not Tamil(NT), 5-Offensive Targeted Insult Other(OTIO)

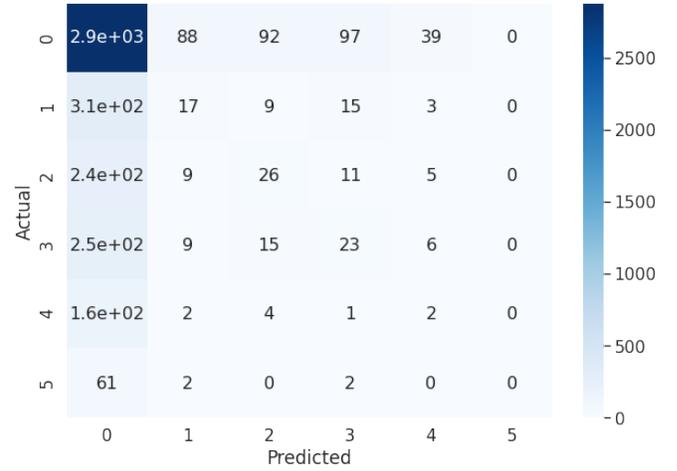

Figure 6: Confusion matrix of the best model - CNN-BiLSTM. 0-Not Offensive(NO), 1-Offensive Untargetede(OU), 2-Offensive Targeted Insult Group(OTIG), 3-Offensive Targeted Insult Individual(OTII), 4-Not Tamil(NT), 5-Offensive Targeted Insult Other(OTIO)

the dataset which has a consequential impact on system performance.

Moreover, in some instances, the STT (1) may give bad translations. For instance, when the sentence is entirely in English, STT algorithm translates every word, affecting the meaning of the sentence because, in general, the word-to-word translation of a sentence will not give the correct meaning in the target language. This is a drawback of our STT algorithm. But if we consider the real-world problem, most of the corpus text is in code-mixed and romanized format reduces the issues from this drawback. Notably, the ULMFiT and mBERT-BiLSTM models perform even better than other transformer-based models (DistilBERT and XLM-RoBERTa) and deep learning-based approach (CNN-BiLSTM).

## 8 Benchmark Systems

We participated in a shared task on offensive language detection in Dravidian languages which was organized as a Codalab competition[10] to encouraging research in under resourced Dravidian languages (Tamil, Malayalam, and Kannada). Many participants connecting to all three languages had submitted their solutions. Table 4 shows the overall results and teams which were placed in the top positions in the benchmark for the Tamil language.

The team **hate-alert** [32] got first position followed by the team **indicnlp@kgp** [33] in the second place. As per the F1 scores, out of 31 teams, our ULMFiT and mBERT-BiLSTM models (indicated as **Current Task** in the Table 4) stand between **OFFLangOne** and **es** teams, lagging the top model only by 0.04 points. This is proven that mBERT along with the BiLSTM reached the state-of-the-art results compared to the benchmark systems and can be also observed that ULMFiT has effectively fine-tuned the language model for our task.

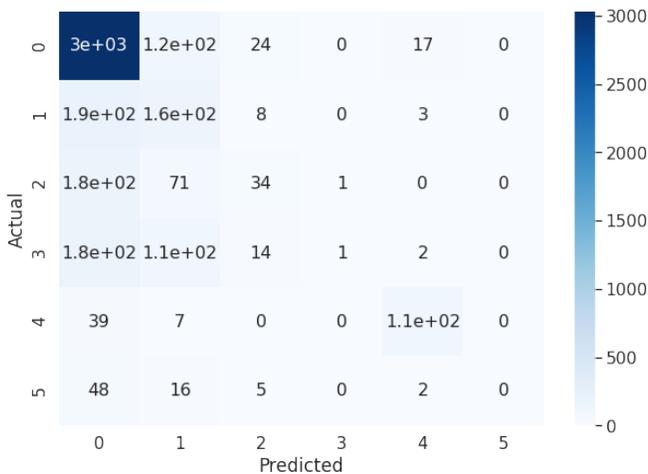

Figure 5: Confusion matrix of the best model - mBERT-BiLSTM. 0-Not Offensive(NO), 1-Offensive Untargetede(OU), 2-Offensive Targeted Insult Group(OTIG), 3-Offensive Targeted Insult Individual(OTII), 4-Not Tamil(NT), 5-Offensive Targeted Insult Other(OTIO)

---

[10] https://competitions.codalab.org/competitions/27654

| Team Name | Precision | Recall | F1-Score |
|---|---|---|---|
| hate-alert [32] | 0.78 | 0.78 | 0.78 |
| indicnlp@kgp [33] | 0.75 | 0.79 | 0.77 |
| ZYJ123 [34] | 0.75 | 0.77 | 0.76 |
| SJ-AJ [35] | 0.75 | 0.79 | 0.76 |
| NLP@CUET [36] | 0.75 | 0.78 | 0.76 |
| Codewithzichao [37] | 0.74 | 0.77 | 0.75 |
| hub [38] | 0.73 | 0.78 | 0.75 |
| MUCS [39] | 0.74 | 0.77 | 0.75 |
| bitions [40] | 0.74 | 0.77 | 0.75 |
| IIITK [41] | 0.74 | 0.77 | 0.75 |
| OFFLangOne [42] | 0.74 | 0.75 | 0.75 |
| **Current Task** | **0.72** | **0.76** | **0.74** |
| cs [43] | 0.74 | 0.75 | 0.74 |
| hypers [23] | 0.71 | 0.76 | 0.73 |
| SSNCSE-NLP [44] | 0.74 | 0.73 | 0.73 |
| JUNLP [45] | 0.71 | 0.74 | 0.72 |
| IIITT [46] | 0.70 | 0.73 | 0.71 |
| IRNLP-DAIICT [47] | 0.72 | 0.77 | 0.71 |
| CUSATNLP [48] | 0.67 | 0.71 | 0.69 |
| Amrita-CEN-NLP [49] | 0.64 | 0.62 | 0.62 |
| JudithJeyafreedaAndrew [50] | 0.54 | 0.73 | 0.61 |

Table 4: Comparison with the benchmark leader board based on F1-score along with other evaluation metrics (Precision and Recall). Our model is indicated as **Current Task**.

## 9 Conclusion

In this work, we have described and analyzed various techniques and neural network models to detect the offensive language in code-mixed romanized social media text in Tamil. A novel technique of selective translation and transliteration is proposed to deal with code-mixed and romanized offensive language classification in Tamil. This technique is flexible and can be extended to other languages as well. For classification, with classical classifiers on top of pre-trained embeddings, we applied and fine-tuned transformer-based models such as mBERT, DistilBERT, and XLM-RoBERTa, which provides an astonishing rise in accuracy than DL-based methods like CNN-BiLSTM Weighted f1 score increased from 0.6112 to 0.7341. We experimented with different transfer learning techniques to leverage the offensive speech dataset from resource-rich languages. Our work also points to the usefulness of Transformer architectures, particularly mBERT, for low resource languages like Tamil. Our proposed models show an average performance of 74% F1-weighted score across the dataset.

## 10 Future Work

In the above work, we implement STT algorithm to translate and transliterate other languages into Tamil native script, even we could reach a bounded results. It may be due to the lower amount of dataset that is used to fine-tune the model and difficulty in finding a Tamil word embedding like GloVe(low-resourced). Since English has the best word embeddings and pre-trained models, we have planned to modify our STT algorithm to translate and transliterate our dataset to English as one of the future works. Since Tamil is a low-resourced language, there are a small number of word embeddings available for it. So, we have decided to create a word embedding for Tamil with consideration of semantic, syntax, and morphology relationships.

Moreover, it is decided to deal with imbalanced data in the training and development sets by applying over-sampling to increase the number of samples for minority classes. It is assumed that the above approaches will increase the model's performance in the future.

## Acknowledgements

We would like to express our thanks to Mr.Sanjeepan Sivapiran[11] and Mr.Temcious Fernando[12] for their helpful suggestions to improve and clarify this manuscript.

---

[11] https://www.linkedin.com/in/sanjeepan/
[12] https://www.linkedin.com/in/temcious-fernando-2a69a0196/

**Code Availability**

All software used in the above article are available from the public data repository at GitHub https://github.com/Chaarangan/ODL-Tamil-SN.

**Conflict of Interest**

On behalf of all authors, the corresponding author states that there is no conflict of interest.